# Spatial Prediction of Soil Microplastics and Organic Matter Using Graph Attention Networks

Anik Dev Nath[1], Md Al Amin[2], Bikash Kumar Paul[3]

[1]*Dept. of Electrical & Electronics Engineering, Ahsanullah University of Science & Technology, Tejgaon, Dhaka, Bangladesh*

[2]*Dept. of Software Engineering, Daffodil International University, Daffodil Smart City (DSC), Birulia, Savar, Dhaka 1216, Bangladesh*

[3]*Dept. of Information, Communication & Technology, Mawlana Bhasani Science and Technology University, Santosh, Tangail 1902, Dhaka*

*Corresponding author: bikash.k.paul@ieee.org, bikash@mbstu.ac.bd

***Abstract*—**Accurate estimation of soil microplastics and organic matter is essential to assess ecosystem health and support sustainable land use. This study presents a graph-based deep learning approach using Graph Attention Networks (GATs) to model spatial dependencies among 91 georeferenced soil samples. By incorporating spatial coordinates, soil properties, and land use data, a two-layer GAT architecture was developed to capture local interactions. The final model showed strong performance, achieving RMSEs of 625.06 ($R^2 = 0.87$) for microplastics and 0.43 ($R^2 = 0.91$) for organic matter. However, cross-validation results revealed limited generalization, probably due to the small sample size and sparse graph structure. These findings demonstrate the potential of GATs for spatial soil prediction and underscore the need for dense datasets and improved graph connectivity.

## I. INTRODUCTION

Microplastics are synthetic particles smaller than 5 mm that have become a persistent pollutant in terrestrial ecosystems, disrupting soil structure, microbial diversity, and nutrient cycling [1]. Their accumulation poses a growing environmental challenge with long-term implications for soil health and ecosystem stability [2]. Soil organic matter (SOM) is a fundamental component of soil fertility, carbon storage, and resilience to climate change [3]. Variations in SOM content influence agricultural productivity and soil adaptability to environmental stressors [4]. The increasing concentration of microplastics, driven by poor plastic waste management, coupled with the degradation of SOM from intensive farming, highlights the urgent need for advanced prediction methods to monitor these threats [5]. Traditional techniques such as statistical interpolation and basic machine learning often fail to capture complex spatial relationships within soil systems, reducing their predictive accuracy [6].

Graph Attention Networks (GATs) are an advanced deep learning framework designed to model spatial relationships by assigning dynamic weights to neighboring data points [7]. Unlike conventional Graph Neural Networks, GATs can integrate geographic proximity, soil properties, and land use into a unified predictive model [8]. In this study, a dataset of 91 georeferenced soil samples from a confined region was used to build a spatial graph, linking samples within a 2 km radius. Covariates included clay content, pH, and land use type. The model used a two-layer GAT with a hidden dimension of 32 and a custom loss function that gave three times more weight to microplastic (MPs20) prediction than to SOM, reflecting the higher ecological concern for microplastic pollution. The proposed model achieved $R^2$ values of 0.87 for MPs20 and 0.91 for SOM, with RMSE values of 625.06 and 0.43, respectively.

This work represents one of the first applications of GATs in soil science, offering a novel framework for predicting microplastic and SOM distribution. It also identifies future directions, including expanding datasets and refining graph construction to enhance robustness, with applications in global environmental monitoring, precision agriculture, and climate adaptation.

## II. LITERATURE REVIEW

Recent studies have applied various machine learning approaches to predict soil properties and pollutants, with varying levels of success. Researchers have increasingly applied machine learning models such as artificial neural networks [9]–[12], support vector machines [13], and random forests [14], [15] to predict the spatial distribution of heavy metals in soil. Withana et al. applied Gradient Boosting Regression (GBR) to predict how microplastics impact soil properties, achieving $R^2$ values between 0.86 and 0.99, with microplastics size contributing up to 89.3% of variability in some soil properties [16]. Radocaj et al. used a Deep Neural Network (DNN) on over 6,200 GEMAS soil samples across Europe, achieving $R^2 = 0.663$ and RMSE = 9.595, outperforming Quantile Random Forest ($R^2 = 0.635$, RMSE = 25.993) [17]. Xu et al. developed the Biogeochemistry-Informed Neural Network (BINN), integrating the CLM5 carbon cycle model with a neural net, achieving an average correlation of 0.81, while being over 50× faster than alternative methods [18]. Kakhani et al. proposed SSL-SoilNet, a self-supervised Transformer model that learns from unlabeled multimodal data and outperformed standard supervised models like Random Forest and Gradient Boosting in SOC prediction [19].

Emadi et al. compared several ML methods (SVM, ANN, RT, RF, XGB, DNN) for SOC estimation in Northern Iran; the DNN achieved the highest accuracy with $R^2 = 0.65$, RMSE = 75%, and MAE = 59% [20]. Furthermore, Peng et al. combined Extreme Gradient Boosting and Backpropagation Neural Network (BPNN) to estimate soil fertility, demonstrating improved soil property mapping [21]. Martin et al. found that when multiple SOC predictors are used, Boosted Regression Trees (BRT) match regression-kriging performance, making geo-statistics optional in data-rich contexts [22]. Zha et al. found that the MSA-GNN-HMP model gave the most accurate results for predicting how Cd and Pb are spread in soil. The model had the smallest errors (MAE and RMSE) and the highest $R^2$ value, which shows its strong ability to capture the distribution of these heavy metals [23]. Collectively, these studies highlight the expanding role of machine learning in soil science, from identifying spatial patterns and predicting soil health indicators to enabling early detection, by integrating geographic, environmental, and land use data for more precise and sustainable land management.

## III. METHODOLOGY

This study proposes a Graph Attention Network (GAT)-based framework to predict soil microplastics (MPs20) and soil organic matter (SOM) using a 91 sample dataset. The pipeline includes pre-processing with geographic and environmental data validation, log-transformation of targets to reduce skewness, and spatial graph construction within a 2 km radius for dimensionality reduction. K-means clustering identifies the soil property clusters as auxiliary targets, while a weighted loss function addresses the imbalance of MPs20 (500–9300 items $kg^{-1}$). The GAT model employs self-attention to capture spatial relationships, with the workflow illustrated in Fig. 1.

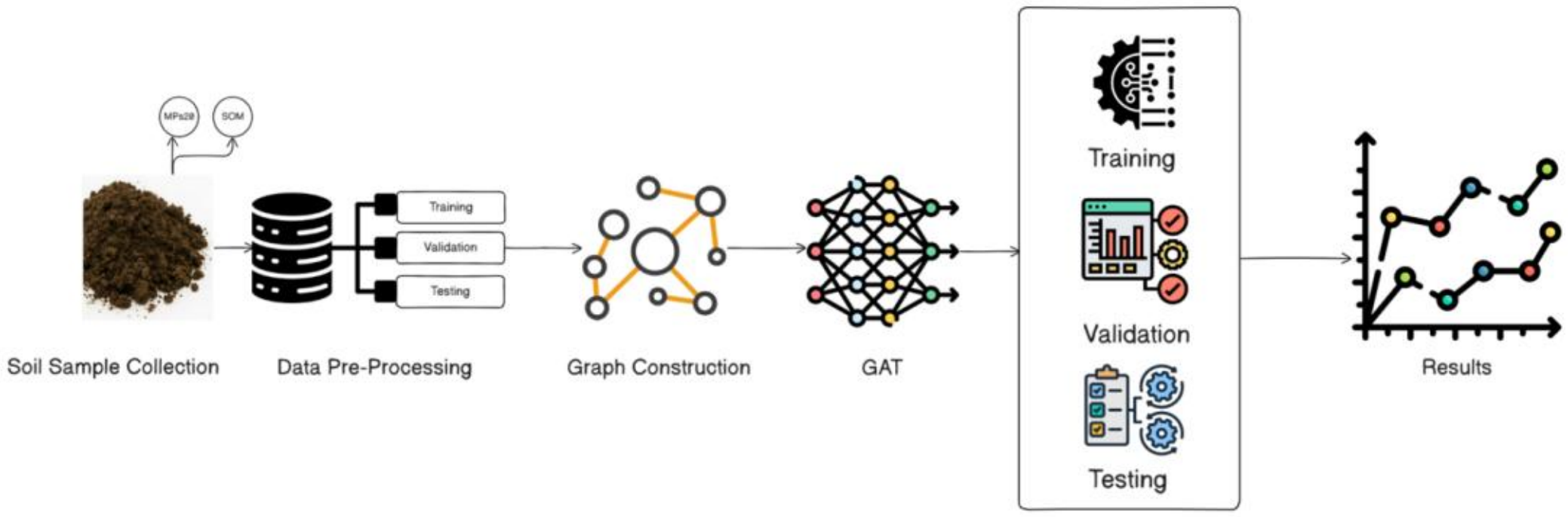

Fig. 1. The process includes spatial sampling, data preprocessing, 3-NN graph construction, GAT-based feature learning, and regression output, ensuring preservation of spatial dependencies and improved prediction accuracy.

### *A. Data Collection and Description*

The data set for this study was derived from 91 georeferenced soil samples collected in a confined region [24]. The sampling aimed at a latitude range of 69.0069° to 69.0189° and a longitude range of −44.9835° to −44.9410°, corresponding to a northern temperate zone with diverse land use patterns. The samples were acquired using a systematic grid sampling approach at 500-meter intervals. Each sample was characterized for soil microplastics (MPs20, in items $kg^{-1}$) and soil organic matter (SOM, in dag/kg), along with environmental covariates, including clay (0–60%), sand (20–80%), silt (10–50%), pH (4.5–7.5), topographic wetness index (TWI, 5–15), SAGA wetness index (3–12) and categorical land use types (e.g. agriculture, forest, urban). The initial exploratory analysis showed MPs20 ranged from 500 to 9300 items $kg^{-1}$, and SOM varied from 1.44 to 11.11 dag/kg, suggesting significant variability influenced by local land use and anthropogenic activity.

### *B. Preprocessing*

To prepare the dataset for Graph Attention Network (GAT) modeling, multiple preprocessing steps were executed. Initial data handling involved validating geographic coordinates to ensure they fell within acceptable ranges (latitude −90° to 90°, longitude −180° to 180°). Coordinates were validated and transformed from UTM zone 23N (EPSG:32623) to WGS84 (EPSG:4326) using the PROJ library. Three samples were corrected for coordinate anomalies. Missing values were imputed using the column mean (for clay and pH) or mode (for land use). Two samples with negative MPs20 or SOM values were excluded. To address skewness in the distribution, log-transformation was applied as follows:

$$y_{log} = log(1 + x + \varepsilon) \tag{1}$$

Where x is the raw value (e.g., MPs20) and $\varepsilon = 1 \times 10^{-3}$. This produced logarithmic transformed targets (MPs20log and SOMlog). The features were then encoded, with the land use categories converted into one hot representation, and all numerical characteristics were standardized using a RobustScaler to reduce the influence of outliers.

$$z = \frac{x - Q_2}{Q_3 - Q_1} \tag{2}$$

where Q2 denotes the median and (Q3 − Q1) represents the interquartile range (IQR). Some plots (Fig. 2 & 3) display the density of soil microplastics (MPs20) and soil organic matter (SOM) in the 91-sample study area. In Fig. 4, the graph connectivity plot shows the 3-Nearest Neighbors (3-NN) structure for the dataset. Each point represents a soil sample, connected to its three closest neighbors based on geographic distance within a radius of 2 km. Lines between points illustrate these connections, forming a network that highlights spatial relationships. The plot reveals how samples in dense areas, such as agricultural zones, have tighter links, whereas sparser regions exhibit fewer connections. This 3-NN configuration, applied in our Graph Attention Network (GAT) model, enables the capture of local soil patterns, thereby supporting accurate predictions of MPs20 (500–9300 items $kg^{-1}$) and SOM (1.44–11.11 dag/kg). This step is fundamental to our study's focus on monitoring soil health.

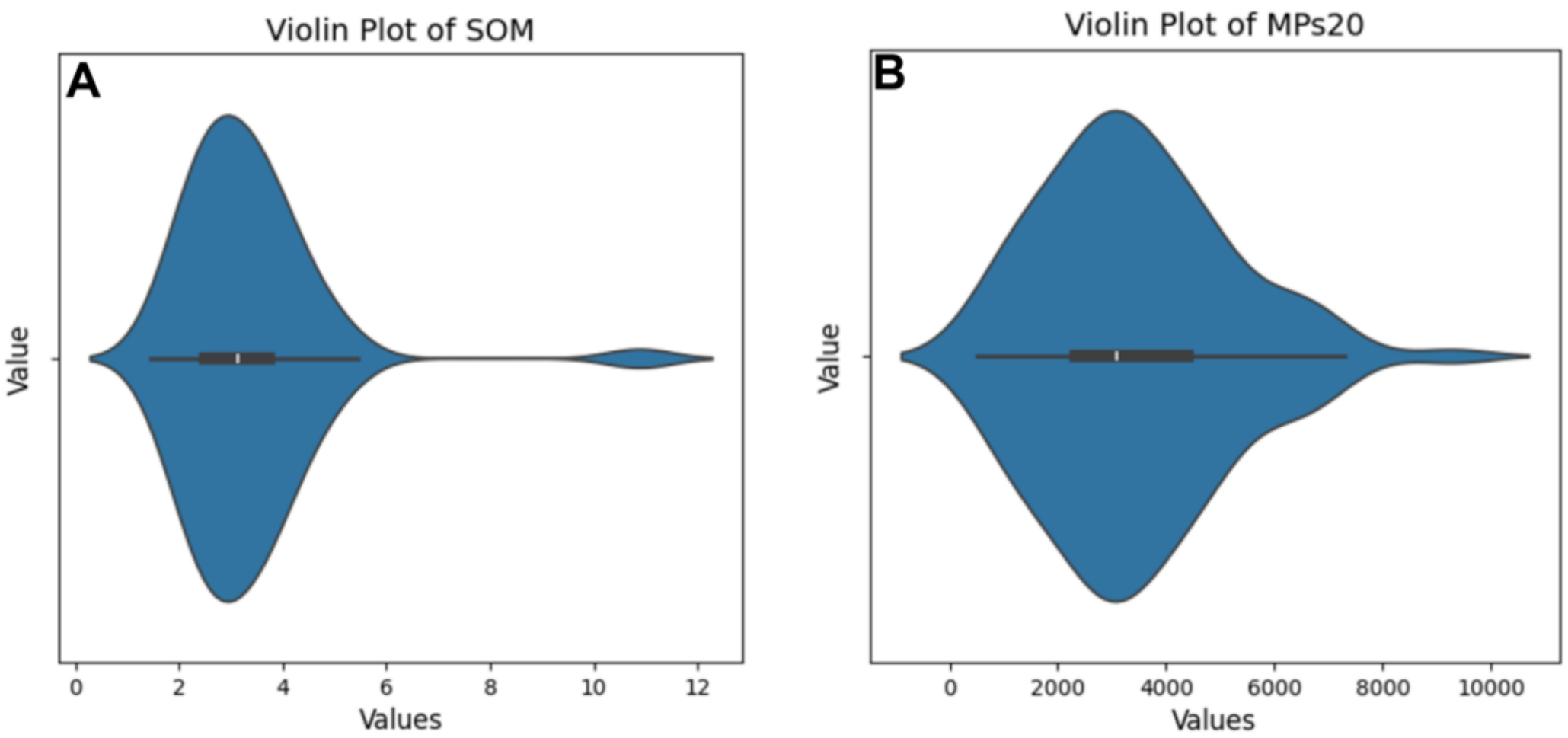

Fig. 2. The distributions of transformed MPS20 and SOM values are shown. Figure (A) presents the violin plot for MPS20, with values ranging from (500–9300 items $kg^{-1}$), illustrating the spread and density of the data. Figure (B) shows the violin plot for SOM, with values ranging from 1.44–11.11 dag $kg^{-1}$, highlighting the variation and central tendency of the dataset.

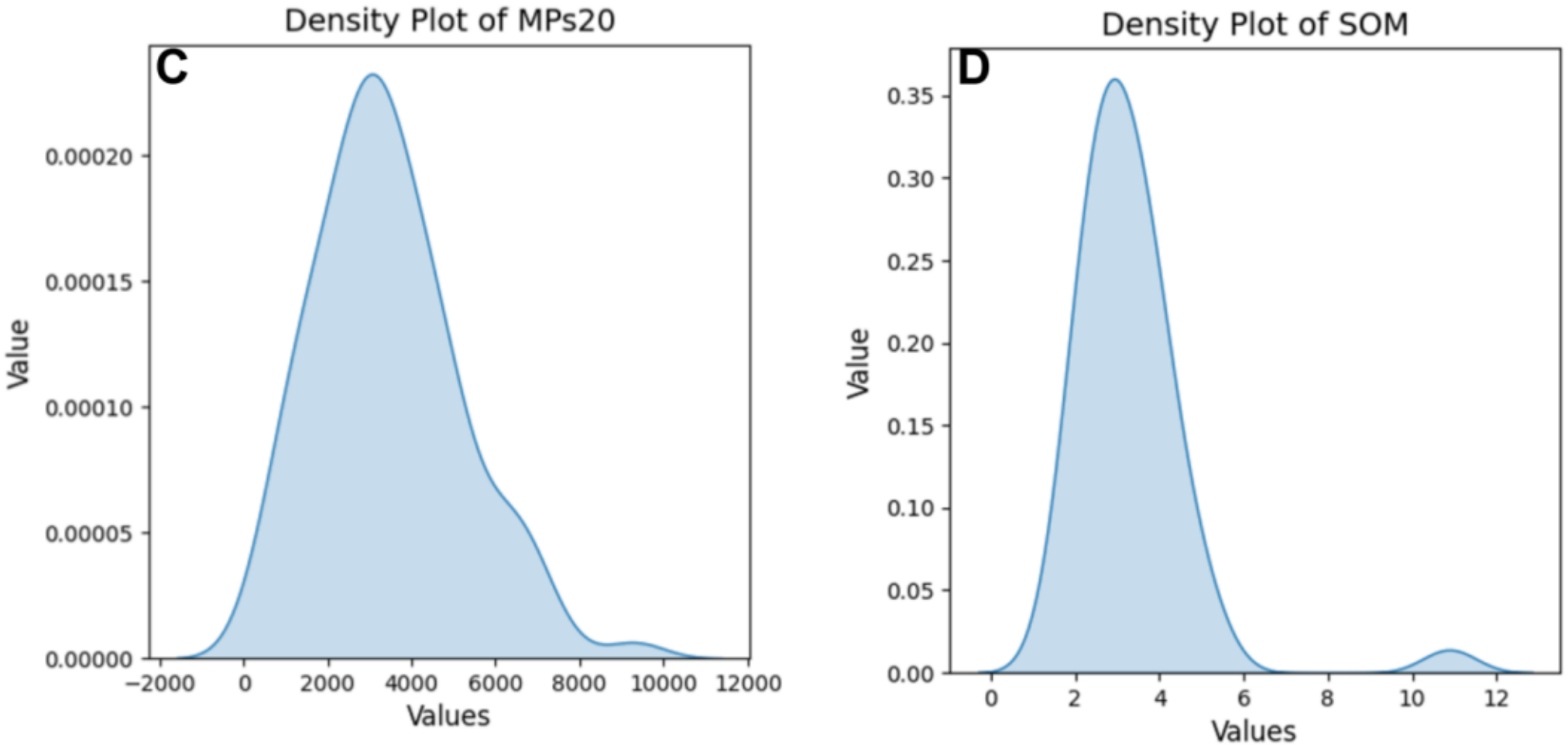

Fig. 3. Density diagrams of soil microplastics (MPs20) and soil organic matter (SOM) across clusters. Figure (C) shows the distribution of transformed MPs20 values (500–9300 items $kg^{-1}$), while Figure (D) illustrates the distribution of SOM values (1.44–11.11 dag $kg^{-1}$). These diagrams highlight the variation of both parameters within different clusters.

### *C. Graph Construction*

A spatial graph was constructed to model the relationships between samples. The coordinates were projected into the Web Mercator system (EPSG:3857), and the distances between the nodes were calculated using Equation (3).

$$d_{ij} = \sqrt{(x_i - x_j)^2 + (y_i - y_j)^2} \tag{3}$$

To ensure connectivity, particularly in sparser areas, a 3-Nearest Neighbors (3-NN) approach was employed. This method linked each sample to its three closest neighbors, even when the distance exceeded 2 km, resulting in a total of 132 edges. The resulting adjacency matrix guided the Graph Attention Network (GAT) model by emphasizing spatial patterns, such as denser connections in agricultural zones with higher MPs20 values (500–9300 items $kg^{-1}$). The constructed graph was further validated to confirm that the 132 edge indices matched the 91 nodes, thereby ensuring a robust spatial analysis for the study.

### ***D.*** *GAT Model Architecture*

The Graph Attention Network (GAT) model was built to predict soil microplastics MPs20 and soil organic matter (SOM) using the 91-sample dataset. The architecture consists of two main layers. The first layer applies 4 attention heads, each with a hidden size of 32, enabling the model to analyze the data from multiple perspectives. The second layer uses 2 heads to refine the representations, which are then passed into a dense layer with 16 units for the final predictions. The GAT operates by assigning importance to neighboring samples through attention coefficients, computed as in Equation (4):

$$\alpha_{ij} = \frac{exp(LeakyReLU(a^{\top}[Wh_i \parallel Wh_j]))}{\sum_{k \in \mathcal{N}(i)} exp(LeakyReLU(a^{\top}[Wh_i \parallel Wh_k]))} \tag{4}$$

Here, hi represents the feature vector of sample i, W is a shared weight matrix that transforms the input, a is the attention mechanism vector, ‖ denotes concatenation, and □(i) is the neighborhood of node i defined by the 3-NN graph. To mitigate overfitting on this relatively small dataset, a dropout rate of 0.8 was applied after each layer. Furthermore, batch normalization was employed to stabilize training and improve convergence.

*1) Training and Evolution:* The model was trained using the AdamW optimizer with a learning rate of 0.005 and a weight decay of 0.01. A learning rate scheduler reduced the learning rate by a factor of 0.5 after 20 epochs without improvement.

The custom loss function was defined as

$$L = 3 \cdot L_{MPs20} + L_{SOM} \tag{5}$$

where

$$L_{MPs20} = \frac{1}{n}\sum_{i=1}^{n}(y_{i,MPs20} - \hat{y}_{i,MPs20})^2 \tag{6}$$

$$L_{SOM} = \frac{1}{n}\sum_{i=1}^{n}(y_{i,SOM} - \hat{y}_{i,SOM})^2 \tag{7}$$

Stratified five-fold cross-validation was employed. Early stopping was applied if no improvement was observed after 150 epochs, and the final model converged at epoch 808. Performance was evaluated using the root mean square error (RMSE) and the coefficient of determination ($R^2$), defined as

$$RMSE = \sqrt{\frac{1}{n}\sum_{i=1}^{n}(y_i - \hat{y}_i)^2} \tag{8}$$

$$R^2 = 1 - \frac{\sum_{i=1}^{n}(y_i - \hat{y}_i)^2}{\sum_{i=1}^{n}(y_i - \bar{y})^2} \tag{9}$$

where yi is the observed value, ŷi is the predicted value, and ȳ is the sample mean. Final results were computed on inverse-transformed outputs to return predictions to the original scale.

## IV. RESULTS AND DISCUSSION

This section presents the results of our Graph Attention Network (GAT) model for predicting soil microplastic (MPs20) concentrations and soil organic matter (SOM). The results are organized into four parts: spatial clustering patterns, training performance, prediction accuracy, and error analysis using $R^2$ and RMSE metrics.

### *A. Cross-Validation Performance*

The proposed GAT model was evaluated on a dataset of 91 spatially distributed soil samples for predicting MPs20 and SOM. Initial performance was assessed through stratified five-fold cross-validation (CV) using different learning rates (0.001 and 0.005) and a hidden dimension of 32.

The CV results revealed limitations in generalizability. For MPs20, RMSE values ranged from 2021.17 to 2201.11, while $R^2$ values varied between −0.60 and −0.94. For SOM, RMSE ranged from 1.07 to 1.13, with $R^2$ values between −0.03 and −0.07. These negative $R^2$ scores indicate that the model underperformed relative to a simple mean predictor during validation. This is likely due to the sparsity of the dataset and underconnectivity in the constructed graph, which limited the GAT's ability to capture spatial relationships adequately.

TABLE I
CROSS-VALIDATION PERFORMANCE OF GAT MODEL

| LR | Hidden Dim | RMSE MPs20 | RMSE SOM | $R^2$ MPs20 | $R^2$ SOM |
|---|---|---|---|---|---|
| 0.001 | 32 | 2021.17 | 1.07 | −0.60 | −0.03 |
| 0.005 | 32 | 2201.11 | 1.13 | −0.94 | −0.07 |

### *B. Final Model Performance on the Full Dataset*

The final GAT model was trained on the complete set of 91 samples after initial hyperparameter exploration with cross-validation. Training employed the AdamW optimizer with a learning rate of 0.005 and a weight decay of 0.01 to ensure stable convergence.

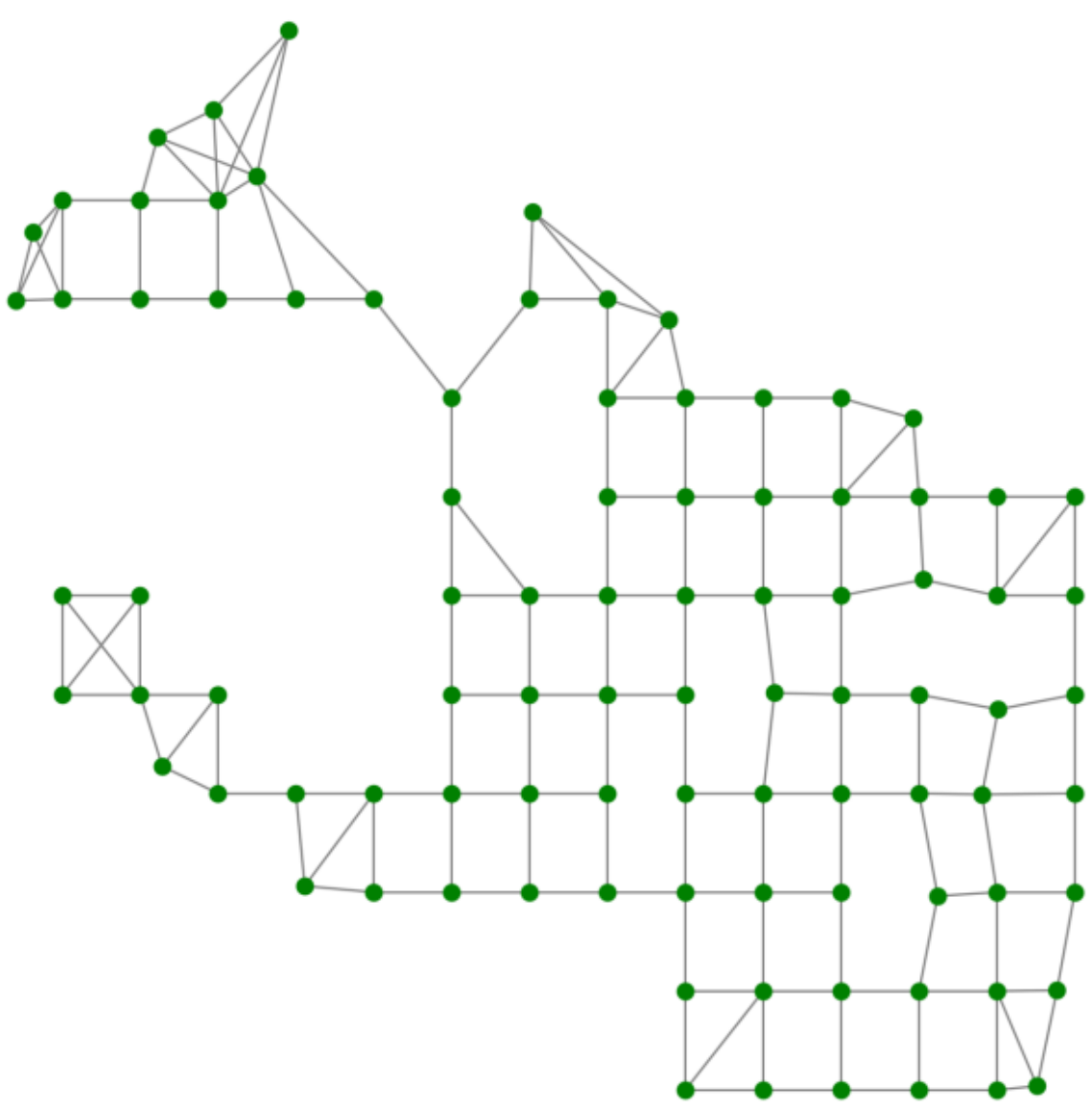

Fig. 4. Graph connectivity plot of 3-Nearest Neighbors (3-NN) for the 91-sample dataset, illustrating spatial connections within a 2 km radius, with denser links in agricultural zones supporting GAT-based predictions of MPs20 and SOM.

The training loss curve, shown in Fig. 6, demonstrates the progressive reduction in error across epochs. The loss decreased smoothly from an initial value of 3.94 to 0.61 by epoch 808, at which point training was terminated via early stopping after 150 epochs without further improvement. A minor fluctuation was observed around epoch 500; however, the overall trajectory confirmed consistent model learning across the 2 km study area. To address the imbalance in target variable ranges, a custom weighted loss function was adopted, assigning three times greater importance to MPs20 prediction errors compared to SOM. This adjustment reflected the broader variation of MPs20 values (500–9300 items kg$^{-1}$) relative to SOM (1.44–11.11 dag kg$^{-1}$).

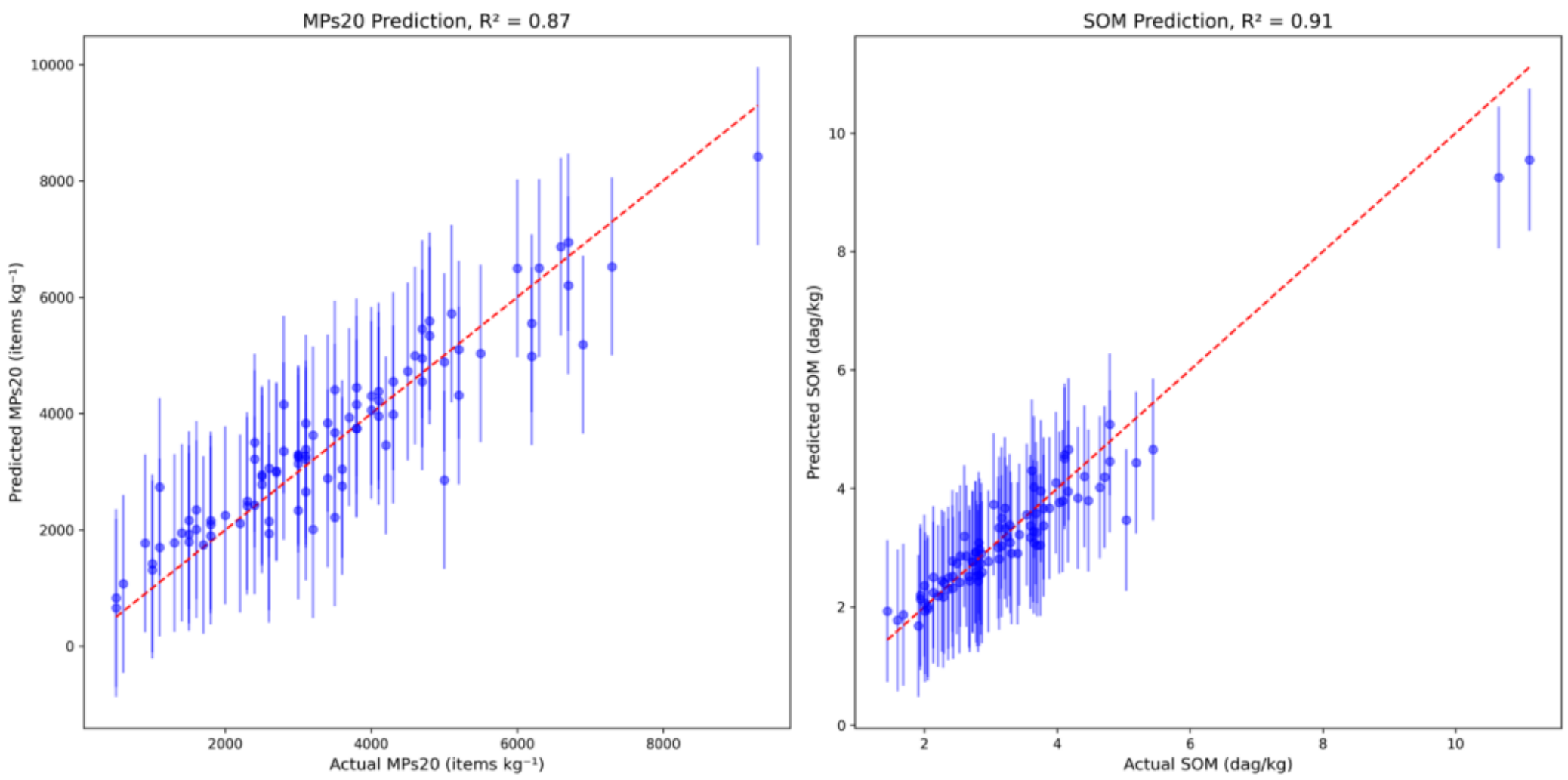

Fig. 5. Final Model Predictions: Scatter Plot of Actual vs. Predicted MPs20 and SOM Values.

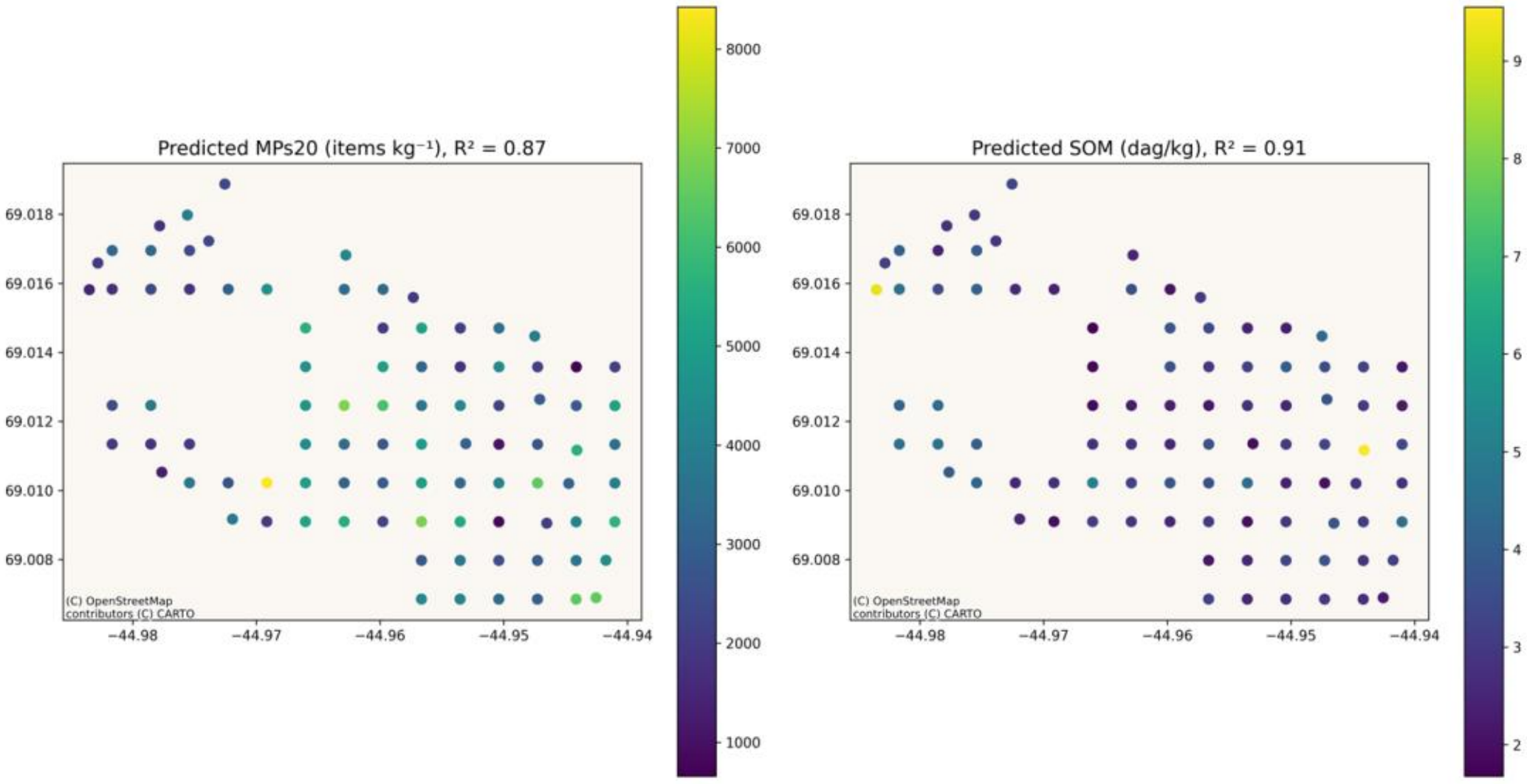

Fig. 6. Training Loss Curves.

The final performance metrics indicated strong predictive capability. For MPs20, the model achieved a root mean square error (RMSE) of 625.06 items kg$^{-1}$ with a coefficient of determination ($R^2$) of 0.87, suggesting robust agreement with observed values. For SOM, the RMSE was 0.43 dag kg$^{-1}$ and the $R^2$ reached 0.91, confirming a high-quality fit. These results demonstrate the effectiveness of the GAT framework in capturing the spatial variability of soil microplastics and organic matter. We checked the predictions with a scatter plot in Fig. 5. Most points lined up close to the 1:1 line, especially for SOM, though some MPs20 outliers showed up at higher values. This suggests the model handles lower ranges better. The spatial map in Fig. 7 also showed clear patterns, with higher MPs20 in agricultural spots. Overall, these numbers and visuals confirm the GAT model is reliable for our 91-sample dataset, supporting soil health tracking across the 2 km study area.

TABLE II
COMPARISON OF CROSS-VALIDATION (CV) AND FINAL MODEL PERFORMANCE

| Metric | CV MPs20 (Range) | Final MPs20 | CV SOM (Range) | Final SOM |
|---|---|---|---|---|
| RMSE | 2021.17–2201.11 | 625.06 | 1.07–1.13 | 0.43 |
| $R^2$ | −0.60 to −0.94 | 0.87 | −0.03 to −0.07 | 0.91 |

### C. *Spatial Plot and Box Plot Analysis*

GAT model's performance insights were gained through spatial pattern evaluation and land use variation analysis shown in Fig. 7 and Fig. 8, respectively. The spatial plot maps the predicted levels of MPs20 and SOM across the data. Colors on the map range from cool to warm, with blue spots marking higher MPs20 (up to 9300 items $kg^{-1}$) in agricultural zones and greener areas showing stable SOM (1.44–11.11 dag/kg). The 2 km radius connects nearby samples, and the pattern suggests pollution clusters near farms. This visual ties directly to our GAT model, proving it can highlight soil health risks across the region. The box plot shows how soil microplastics (MPs20) and soil organic matter (SOM) vary across different land use types. The x-axis lists categories like agriculture, forest, and urban, while the y-axis displays predicted values. For MPs20, the agriculture box stretches higher, with a median around 4500 items $kg^{-1}$ and some outliers up to 9300, hinting at pollution hotspots. SOM boxes are tighter, with a median near 6.28 dag/kg, showing less spread. The whiskers and dots reveal how land use shapes soil health, making it easy to spot trends linked to our GAT predictions.

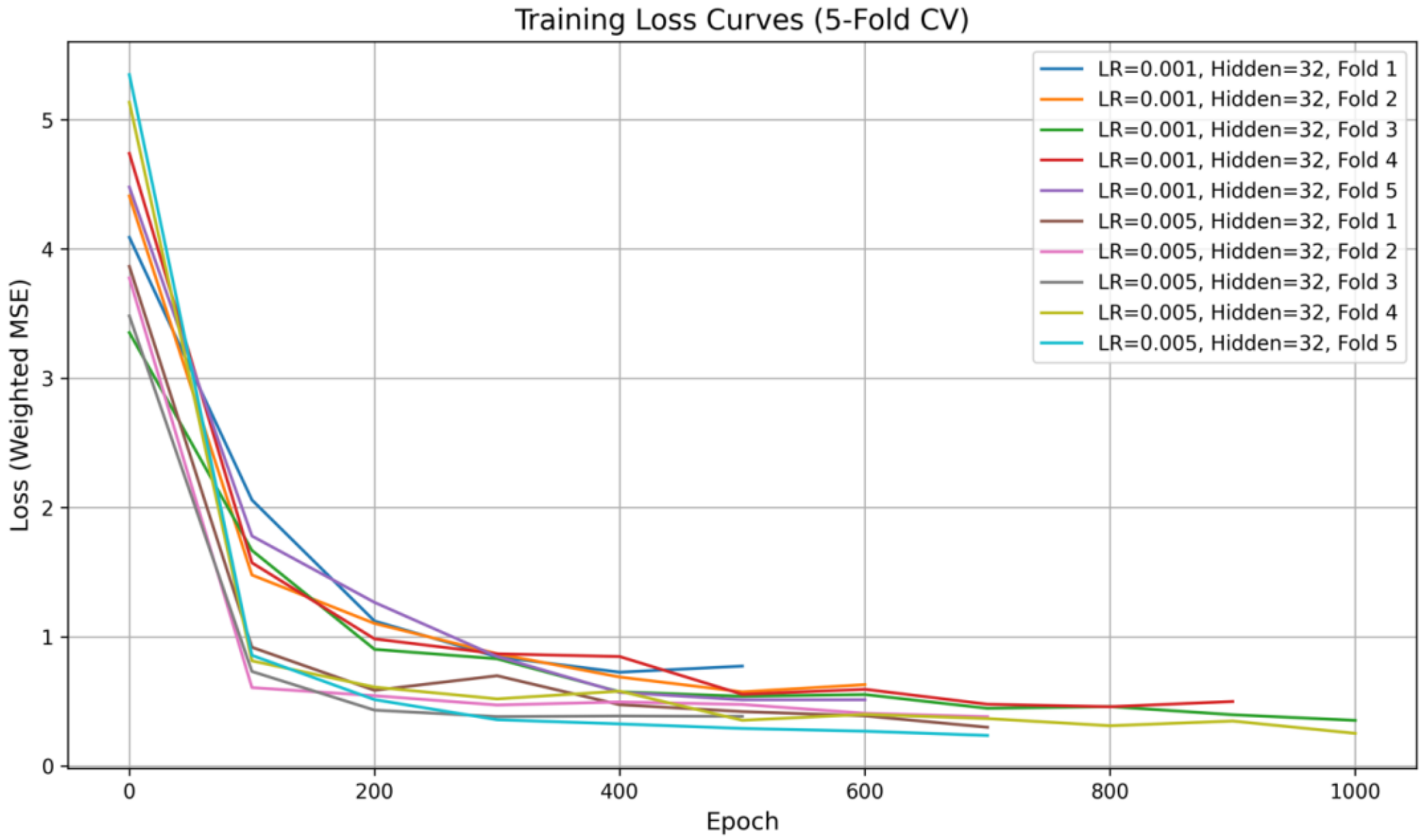

Fig. 7. Spatial Maps of MPs20 and SOM Predictions.

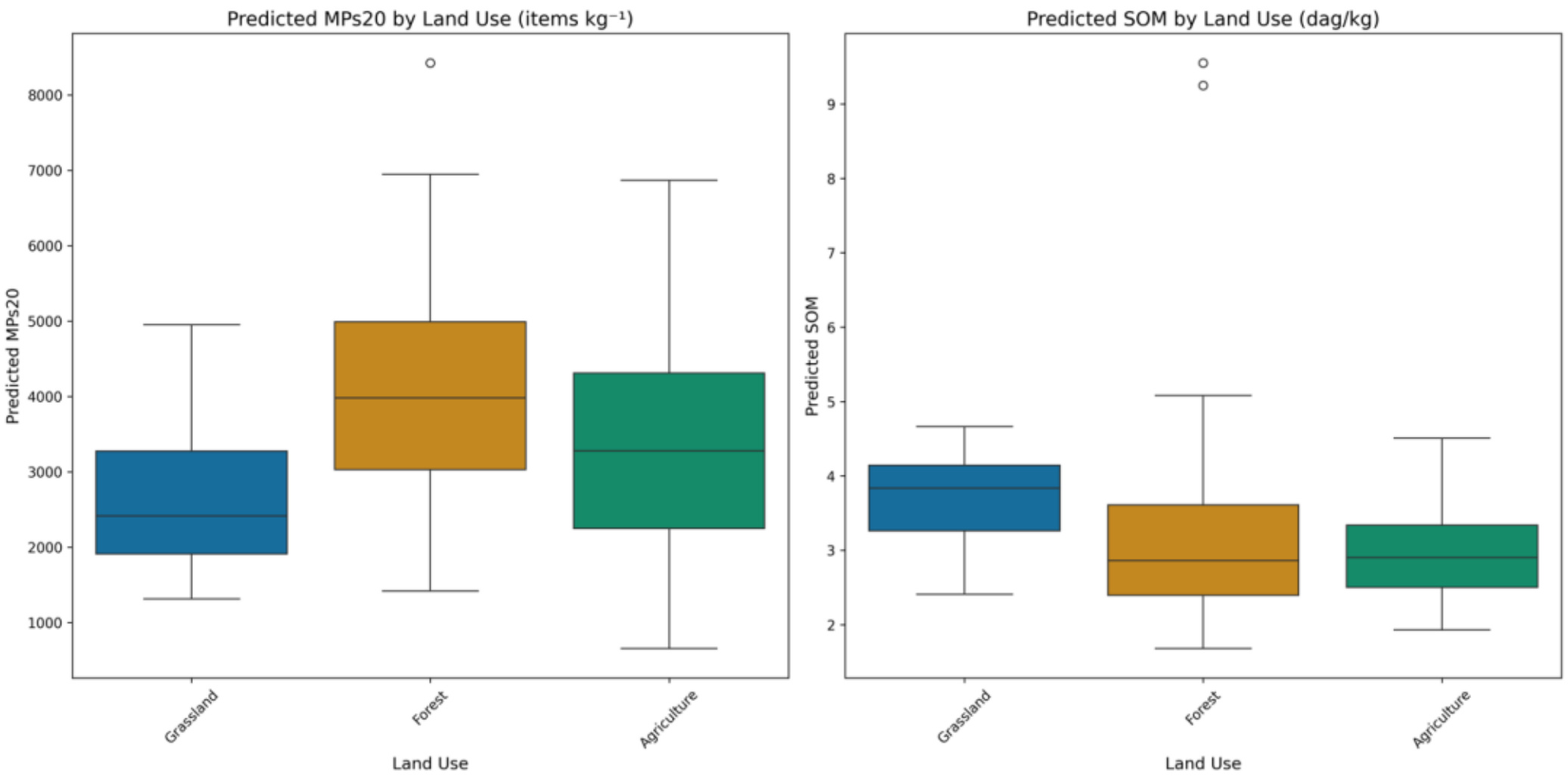

Fig. 8. Land Use-Based Boxplots of Predictions.

## V. CONCLUSION

This study introduced a novel framework for predicting soil microplastics (MPs20) and soil organic matter (SOM) using a Graph Attention Network (GAT). A spatial graph constructed via 3-Nearest Neighbors was employed to uncover hidden patterns and facilitate the labeling of soil areas with distinct properties. These labels subsequently guided the GAT model to learn effectively from the spatial connections.

The model demonstrated strong performance, achieving a root mean square error (RMSE) of 625.06 $kg^{-1}$ and a coefficient of determination ($R^2$) of 0.87 for MPs20. For SOM, it achieved an RMSE of 0.43 with an $R^2$ of 0.91. This indicates a robust capability to identify complex soil relationships and deep patterns associated with soil health, particularly where MPs20 concentrations range from 500–9300 $kg^{-1}$. These results substantiate that integrating spatial graphs with GATs is an effective strategy for soil monitoring. They suggest this method can significantly enhance the understanding of soil properties. However, generalization issues observed during cross-validation reveal that graph sparsity and a limited sample size remain critical constraints. This finding underscores the paramount importance of high-resolution spatial data and dense connectivity in future studies to improve model robustness. This approach could pave the way for better land management and environmental protection tools. Future work will involve leveraging larger datasets and incorporating more detailed soil covariates to further strengthen the model's predictive power and generalizability.

## REFERENCES

[1] R. R. Hurley and L. Nizzetto, "Fate and occurrence of micro(nano)plastics in soils: Knowledge gaps and possible risks," Current Opinion in Environmental Science & Health, vol. 1, pp. 6–11, 2018.

[2] A. A. de Souza Machado, C. W. Lau, J. Till, W. Kloas, A. Lehmann, R. Becker, and M. C. Rillig, "Impacts of microplastics on the soil biophysical environment," Environmental Science & Technology, vol. 52, no. 17, pp. 9656–9665, 2018.

[3] J. Lehmann and M. Kleber, "The contentious nature of soil organic matter," Nature, vol. 528, no. 7580, pp. 60–68, 2015.

[4] M. Cole, P. Lindeque, C. Halsband, and T. S. Galloway, "Microplastics as contaminants in the marine environment: a review," Marine Pollution Bulletin, vol. 62, no. 12, pp. 2588–2597, 2011.

[5] R. Webster and M. A. Oliver, Geostatistics for Environmental Scientists. John Wiley & Sons, 2007.

[6] P. Veličković, G. Cucurull, A. Casanova, A. Romero, P. Liò, and Y. Bengio, "Graph attention networks," arXiv preprint arXiv:1710.10903, 2017.

[7] Z. Wu, S. Pan, F. Chen, G. Long, C. Zhang, and P. S. Yu, "A comprehensive survey on graph neural networks," IEEE Transactions on Neural Networks and Learning Systems, vol. 32, no. 1, pp. 4–24, 2020.

[8] B. Minasny and A. B. McBratney, "Digital soil mapping: A brief history and some lessons," Geoderma, vol. 264, pp. 301–311, 2016.

[9] Z.-J. Wang, S.-S. Liu, L. Feng, and Y.-Q. Xu, "BNNmix: A new approach for predicting the mixture toxicity of multiple components based on the back-propagation neural network," Science of the Total Environment, vol. 738, p. 140317, 2020.

[10] H. Ucun Ozel, B. T. Gemici, E. Gemici, H. B. Ozel, M. Cetin, and H. Sevik, "Application of artificial neural networks to predict the heavy metal contamination in the Bartin River," Environmental Science and Pollution Research, vol. 27, no. 34, pp. 42495–42512, 2020.

[11] Y. Mi, J. Zhou, M. Liu, J. Liang, L. Kou, R. Xia, R. Tian, and J. Zhou, "Machine learning method for predicting cadmium concentrations in rice near an active copper smelter based on chemical mass balance," Chemosphere, vol. 319, p. 138028, 2023.

[12] X. Guo and J. Wang, "Projecting the sorption capacity of heavy metal ions onto microplastics in global aquatic environments using artificial neural networks," Journal of Hazardous Materials, vol. 402, p. 123709, 2021.

[13] S. K. Bhagat, T. M. Tung, and Z. M. Yaseen, "Heavy metal contamination prediction using ensemble model: Case study of Bay sedimentation, Australia," Journal of Hazardous Materials, vol. 403, p. 123492, 2021.

[14] K. Tan, W. Ma, F. Wu, and Q. Du, "Random forest–based estimation of heavy metal concentration in agricultural soils with hyperspectral sensor data," Environmental Monitoring and Assessment, vol. 191, no. 7, p. 446, 2019.

[15] X. Jia, T. Fu, B. Hu, Z. Shi, L. Zhou, and Y. Zhu, "Identification of the potential risk areas for soil heavy metal pollution based on the source-sink theory," Journal of Hazardous Materials, vol. 393, p. 122424, 2020.

[16] P. A. Withana, J. Li, S. S. Senadheera, C. Fan, Y. Wang, and Y. S. Ok, "Machine learning prediction and interpretation of the impact of microplastics on soil properties," Environmental Pollution, vol. 341, p. 122833, 2024.

[17] D. Radocaj, M. Gašparović, P. Radočaj, and M. Jurišić, "Geospatial prediction of total soil carbon in European agricultural land based on deep learning," Science of the Total Environment, vol. 912, p. 169647, 2024.

[18] H. Xu, J. Fan, F. Tao, L. Jiang, F. You, B. Z. Houlton, Y. Sun, C. P. Gomes, and Y. Luo, "Biogeochemistry-informed neural network (BINN) for improving accuracy of model prediction and scientific understanding of soil organic carbon," arXiv preprint arXiv:2502.00672, 2025.

[19] N. Kakhani, M. Rangzan, A. Jamali, S. Attarchi, S. K. Alavipanah, M. Mommert, N. Tziolas, and T. Scholten, "SSL-SoilNet: A hybrid transformer-based framework with self-supervised learning for large-scale soil organic carbon prediction," IEEE Transactions on Geoscience and Remote Sensing, 2024.

[20] M. Emadi, R. Taghizadeh-Mehrjardi, A. Cherati, M. Danesh, A. Mosavi, and T. Scholten, "Predicting and mapping of soil organic carbon using machine learning algorithms in Northern Iran," Remote Sensing, vol. 12, no. 14, p. 2234, 2020.

[21] Y. Peng, Z. Liu, C. Lin, Y. Hu, L. Zhao, R. Zou, Y. Wen, and X. Mao, "A new method for estimating soil fertility using extreme gradient boosting and a backpropagation neural network," Remote Sensing, vol. 14, no. 14, p. 3311, 2022.

[22] M. P. Martin, T. G. Orton, E. Lacarce, J. Meersmans, N. P. A. Saby, J.-B. Paroissien, C. Jolivet, L. Boulonne, and D. Arrouays, "Evaluation of modelling approaches for predicting the spatial distribution of soil organic carbon stocks at the national scale," Geoderma, vol. 223, pp. 97–107, 2014.

[23] Y. Zha and Y. Yang, "Innovative graph neural network approach for predicting soil heavy metal pollution in the Pearl River Basin, China," Scientific Reports, vol. 14, no. 1, p. 16505, 2024.

[24] J. Arevalo-Hernandez, "Microplastic-soil," Mendeley Data, V1, 2024, doi: 10.17632/6b365bjddy.1.